# AF-Mamba: Efficient Long-Term Signal Modeling for Early Prediction of Atrial Fibrillation Onset

Yongbin Lee[1] and Ki H. Chon[1*], *Fellow, IEEE*

***Abstract*— Atrial fibrillation (AF) is the most common cardiac arrhythmia and is associated with increased risks of stroke and heart failure. The growing availability of wearable and portable ECG monitoring enables continuous assessment of cardiac rhythm outside clinical settings. Predicting AF before its onset could provide additional lead time for timely clinical assessment and potentially improve the management of patients at risk of AF-related complications. This study focuses on predicting AF onset one hour in advance using long-term RR intervals (RRIs). To address this challenge, we propose a deep learning architecture that integrates temporal convolutional networks (TCNs) for local features encoding with Mamba, a selective state-space model capable of long-range sequence modeling. This hybrid TCN-Mamba design enables efficient training and inference on one-hour input windows, overcoming limitations of Transformers' quadratic scaling and recurrent networks' vanishing gradients. In subject-wise 5-fold testing, the proposed model achieved a sensitivity of 0.889, specificity of 0.943, F1-score of 0.813, AUROC of 0.974, and AUPRC of 0.933. In paired cross-dataset holdout evaluation, AF-Mamba maintained discriminative performance across unseen AF and NSR datasets, achieving a mean AUROC of 0.897. Compared against state-of-the-art AF prediction models and general time-series models, AF-Mamba achieved competitive predictive performance while providing a favorable performance–efficiency trade-off for long RRI sequences. These findings demonstrate the potential of AF-Mamba for accurate AF prediction one hour in advance and real-time continuous ambulatory monitoring.**

## I. Introduction

ATRIAL fibrillation (AF) is the most common cardiac arrhythmia and is associated with increased risks of stroke, heart failure, and other cardiovascular complications [1]. Beyond cardiovascular diseases (CVDs), AF is also a major risk factor for cognitive decline and dementia, highlighting its broad impact on long-term health [2].

While AF is a chronic management challenge, a substantial proportion of patients remain asymptomatic, with asymptomatic AF accounting for up to 40% of diagnosed cases [3]. Asymptomatic patients may be less likely to receive timely therapy and have been shown to exhibit a higher risk of disease progression, including a 43% greater hazard of progression to permanent AF compared with symptomatic patients [4]. The growing availability of wearable and ambulatory cardiac monitoring technologies has increasingly enabled continuous cardiac and physiological monitoring outside clinical settings [5], [6], [7]. However, most existing ambulatory monitoring approaches primarily focus on detecting AF after its onset [5], [8], [9]. Predicting AF before its onset could provide clinically meaningful lead time for closer monitoring and timely assessment. Therefore, extending ambulatory monitoring beyond AF detection toward early prediction may improve its utility for long-term AF management. Accordingly, there is a need for predictive approaches that can leverage long-duration ambulatory cardiac signals to identify impending AF before onset.

Several approaches have been proposed to address this challenge, progressing from feature-based machine learning to complex deep learning architectures. Early studies by Bashar et al. and Guo et al. relied on traditional machine learning (ML) techniques utilizing heart rate variability (HRV) features [7], [10], while subsequent works by Gavidia et al. leveraged convolutional neural networks (CNNs) to automate feature extraction [11]. More recently, Li et al. introduced a self-supervised learning approach utilizing Transformer-based architectures to enhance feature representation [12], whereas Gregoire et al. demonstrated that ensemble ML methods could effectively handle long-term (1-hour) inputs to extend the prediction horizon [13].

Despite these advances, significant limitations remain. Traditional ML and CNNs often struggle to capture long-range temporal dependencies due to reliance on extracted features or limited receptive fields. Conversely, while recurrent neural networks (RNNs) and Transformers address sequence modeling, RNNs suffer from the vanishing gradient problem, and Transformers suffer from the quadratic complexity of computational time. Therefore, we propose a deep learning model based on a temporal convolutional network (TCN) and a selective state space model known as Mamba [14], [15]. TCN employs causal convolutions to prevent information leakage from the future and uses dilated convolutions that exponentially expand the receptive field. This design facilitates the extraction of local temporal patterns associated with pre-AF status.

Complementing the TCN, Mamba is a selective state space model that has recently demonstrated strong performance in time-series applications, particularly for long-term inputs and lightweight modeling [16]. Unlike Transformers, which suffer from quadratic time complexity, and RNNs, which are limited by sequential computation and vanishing gradients, Mamba enables efficient parallel training with linear time complexity while preserving robust sequence modeling capacity. Consequently, Mamba provides an efficient framework for modeling long-range temporal patterns associated with pre-AF status from long-duration RRI inputs.

[1]Yongbin Lee, Ki H. Chon are with the Department of Biomedical Engineering, University of Connecticut, Storrs, CT 06269, USA
*Corresponding author: Ki H. Chon (e-mail: ki.chon@uconn.edu).

In this study, we combine TCN for short-term local feature extraction with Mamba for efficient long-range temporal modeling to predict AF using long-duration inputs. We evaluate the model on five public long-term Holter/ambulatory RRI datasets. To assess model performance, we compare our approach with prior models developed specifically for AF prediction, as well as widely used state-of-the-art time-series classification models. Our goal is to develop a robust, accurate, and computationally efficient model capable of predicting AF one hour in advance from long-duration RRI inputs.

## II. Problem Description

The upper and lower panels of Fig. 1 represent 2-hour segments of RR interval (RRI) signals from both AF and normal sinus rhythm (lower panel) subjects, respectively. RRIs represent the time duration between consecutive R-peaks in electrocardiogram (ECG) signals. Because RRI reflects beat-to-beat variability and autonomic regulation, it provides a compact and noise-robust representation of cardiac dynamics suitable for early AF prediction.

Fig. 1(a) shows RRI recordings from 2 hours before the onset of AF to the onset itself. Thirty minutes prior to AF onset (highlighted in red), the RRI segment shows an increase in ectopic beats, including premature atrial contractions (PACs) or premature ventricular contractions (PVCs). However, from 2 hours to 1 hour before AF onset (highlighted in purple), the RRI segment shows fewer PACs/PVCs. Fig. 1(b) shows a 2-hour RRI segment from NSR dataset (no AF) collected from a clinical cardiac monitoring environment. Importantly, irregular RRI patterns can also occur in NSR segments (highlighted in green) that resemble pre-AF activity. Notably, the early pre-AF segment (purple) exhibits fewer ectopic beats than the NSR segment (green), illustrating that ectopic activity alone may not be sufficient to distinguish pre-AF from NSR patterns. To address this difficulty, we used 1-h RRI segments from 2 h to 1 h before AF onset as model inputs, corresponding to a 1-h prediction horizon.

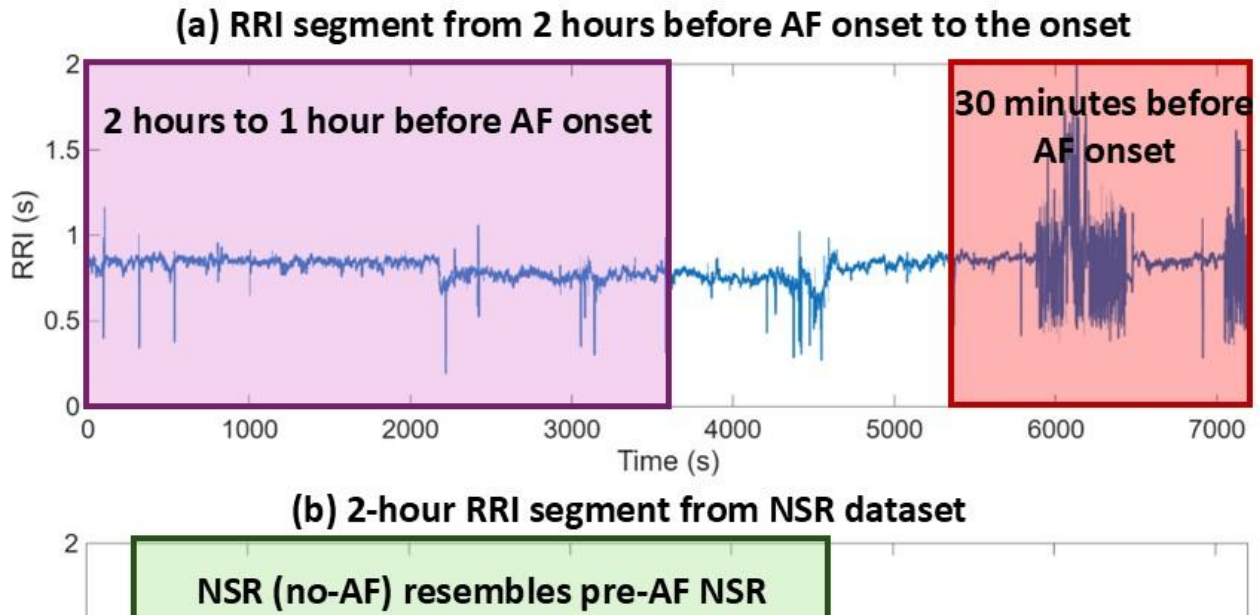

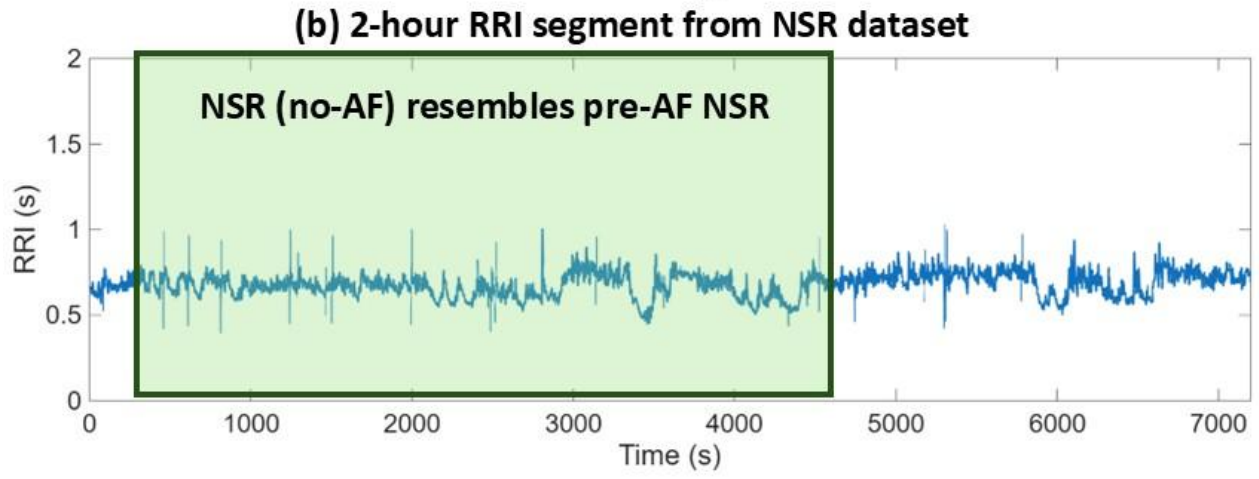

Fig. 1. Two-hour RRI segments. (a) A RRI segment from the AF dataset: more ectopic beats are observed in the final 30 minutes before AF onset. (b) A RRI segment from the NSR dataset: irregular RRI patterns can appear, resembling a pre-AF RRI segment.

## III. Dataset

### A. Dataset Sources and Cohort Selection

In this study, we used five publicly available long-term RRI datasets obtained from Holter or ambulatory cardiac recordings. IRIDIA-AF dataset contains long-term recordings of paroxysmal AF (PAF) obtained from outpatient monitoring, comprising 152 subjects with 19–95 hours of data [17]. LTAF dataset includes long-term recordings of both sustained and paroxysmal AF, consisting of 84 subjects with approximately 24-hour recordings [18]. The MIT-BIH AF dataset provides about 10-hour recordings containing both sustained and paroxysmal AF from 25 subjects [19]. For non-AF controls, we used the MIT-BIH Normal Sinus Rhythm (NSR) dataset and

TABLE I
Demographic and Clinical Characteristics of the Datasets Used in This Study

| Dataset | Class | Source N | Included N | Age, years | Sex (M/F) | Available clinical characteristics | Setting/duration |
|---|---|---|---|---|---|---|---|
| IRIDIA-AF | AF | 152 | 137 | 72 ± 11 (41 – 99) | 81/71 | Paroxysmal AF; mean $CHA_2DS_2$-VASc 3.16 | Outpatient cardiology; 2-channel Holter ECG (fs = 200 Hz), 19 – 95 h |
| LTAF | AF | 84 | 9 | NR | NR | Paroxysmal or sustained AF; multiple annotated rhythms (e.g., SVTA, VT) | 2-channel Holter ECG (fs = 128 Hz), 24 – 25 h |
| MIT-BIH AF | AF | 25 | 14 | NR | NR | Mostly paroxysmal AF; AF, AFL, and AV junctional rhythm annotations | 2-channel ambulatory ECG (fs = 250 Hz), 10 h |
| MIT-BIH NSR | NSR | 18 | 18 | Men: 26 – 45; Women: 20 – 50 | 5/13 | Subjects referred to an arrhythmia laboratory; no significant arrhythmias | 2-channel ambulatory ECG (fs = 128 Hz), 24 h |
| NSR-RR | NSR | 54 | 54 | Men: 28.5 – 76; Women: 58 – 73 | 30/24 | NSR; detailed individual characteristics not provided | RRI annotations from 24-h long-term ECG (fs = 128 Hz); ECG unavailable |

*NR: not reported in the public dataset

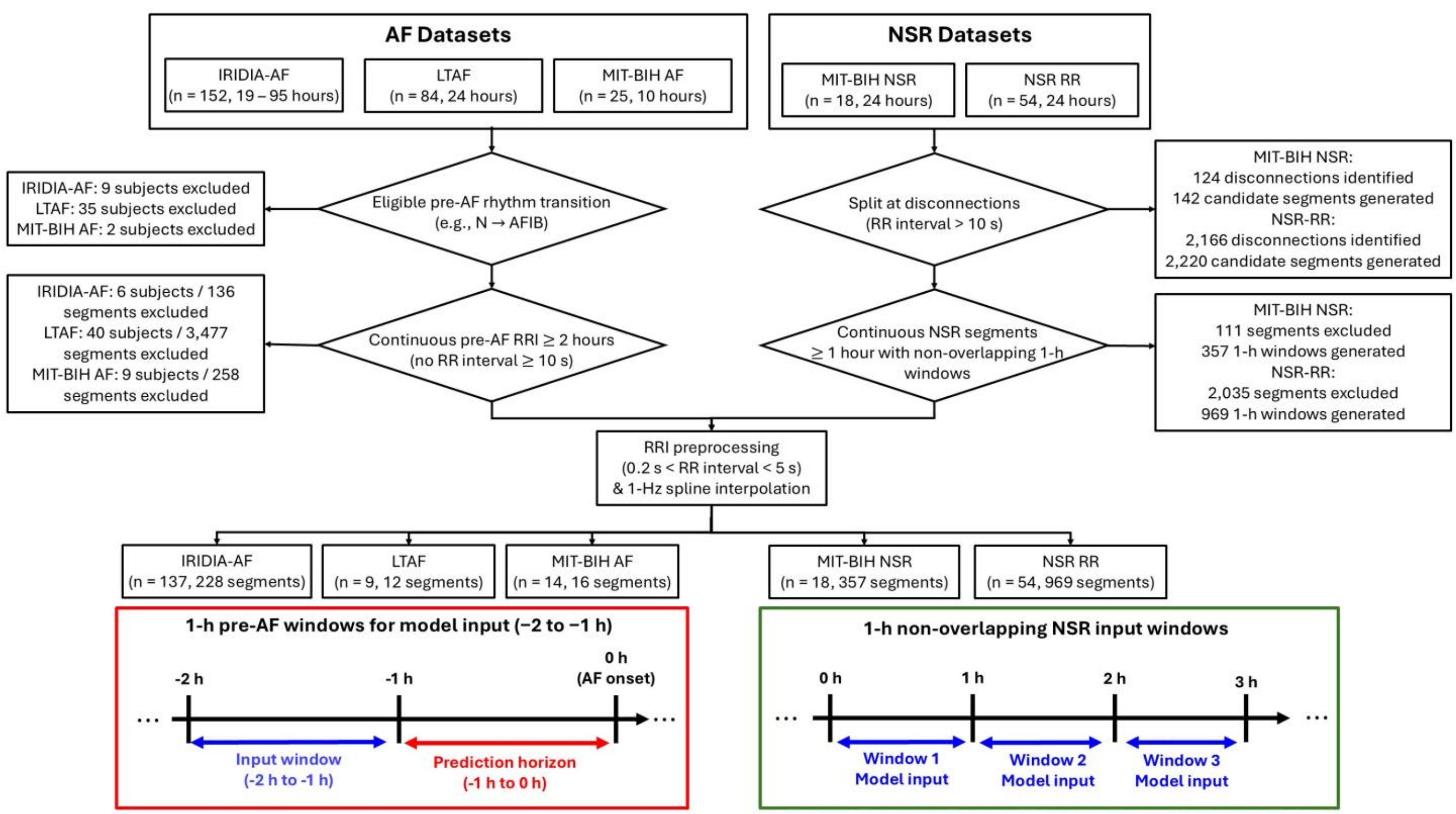

Fig. 2. Flow diagram of subject and segment selection and window generation for AF and NSR datasets.

the NSR RR Interval dataset (NSR RR), containing long-term (approximately 24-hour) NSR recordings of non-AF subjects from 18 and 54 subjects, respectively [20]. The available demographic, rhythm, and recording characteristics of each dataset are summarized in Table I.

For the AF datasets, subjects were eligible when an identifiable AF onset was preceded by an eligible NSR rhythm and at least 2 hours of continuous pre-AF RRI data were available. In the IRIDIA-AF dataset, 137 of 152 subjects met the final inclusion criteria. In the LTAF dataset, 35 of 84 subjects were excluded because AF onset was not preceded by an eligible NSR rhythm, and an additional 40 subjects were excluded because at least 2 hours of continuous pre-AF RRI data were unavailable, resulting in 9 included subjects. In the MIT-BIH AF dataset, 14 of 25 subjects met the final inclusion criteria. Overall, the AF cohort consisted of 160 subjects, while all 18 MIT-BIH NSR subjects and 54 NSR-RR subjects were retained as the NSR cohort, resulting in 72 NSR subjects. The overall subject and segment selection procedure is summarized in Fig. 2.

### B. Window Extraction and Data Splitting

For the AF datasets, 1-hour RRI segments from 2 hours to 1 hour before AF onset were used as model inputs, corresponding to a 1-hour prediction horizon. For the NSR datasets, recordings were split at discontinuities defined by RR intervals greater than 10 s, and continuous segments shorter than 1 hour were excluded. The remaining NSR segments were divided into non-overlapping 1-hour windows. RR intervals outside the physiologically plausible range of 0.2–5 s were removed, and the RRI signals were interpolated at 1 Hz using cubic spline interpolation.

All training, validation, and testing samples were generated using strictly non-overlapping 1-hour windows. This resulted in 256 pre-AF windows from 160 AF subjects and 1,326 NSR windows from 72 NSR subjects. Specifically, 228, 12, and 16 pre-AF windows were obtained from IRIDIA-AF, LTAF, and MIT-BIH AF, respectively, whereas 357 and 969 NSR windows were obtained from MIT-BIH NSR and NSR-RR, respectively.

For model development, AF and NSR subjects were independently divided into five mutually exclusive groups for subject-wise 5-fold cross-validation. In each fold, three groups (60%) were used for training, one group (20%) for validation, and one group (20%) for testing. All windows from the same subject were assigned exclusively to a single split, ensuring strict separation of subjects across the training, validation, and test sets.

## IV. Method

### A. Pre-processing

Before using RRI segments as model inputs, preprocessing steps were performed. First, if any RR interval exceeded 10 seconds (potentially representing extreme bradycardia or signal artifacts), the signal was split at that point without imputation, and the resulting continuous segments were processed separately. RRI segments shorter than 1 hour were discarded because a complete 1-hour model input could not be constructed. Within each segment, RR intervals shorter than 0.2 seconds or longer than 5 seconds were removed. These preprocessing steps were performed to eliminate physiologically implausible values and noise artifacts. The remaining RRI signals were interpolated at 1 Hz using cubic spline interpolation. As all R-peak locations are provided in the source datasets, RR intervals were computed directly from the annotations without applying additional peak-detection algorithms. For the LTAF and MIT-BIH AF datasets, all rhythm annotations other than NSR and AF (e.g., atrial flutter, ventricular tachycardia, etc.) were removed; only NSR segments occurring before AF onset were used for model development.

Fig. 3. Overall architecture of the proposed AF-Mamba model for early atrial fibrillation prediction. The model consists of three main components: (a) TCN blocks extract local temporal features using residual dilated causal convolutions; (b) a long-range modeling based on the Mamba, combined with FFN; and (c) a prediction head that aggregates global features through GAP and GMP, followed by FC layers to classify pre-AF or NSR windows.

### B. Local Feature Extraction with Residual TCN Blocks

Fig. 3 summarizes the architecture of the proposed AF prediction model. Fig. 3(a) illustrates the local temporal feature extraction stage based on TCN blocks. The input RRI sequence of size (1 × 3600) is first processed by three stacked residual TCN blocks. Each block contains two causal dilated 1D convolutions (kernel size = 3, stride = 1), followed by Batch Normalization (BN), ReLU activation, and dropout (rate = 0.2). A residual 1 × 1 Conv1D shortcut is applied in the first block to match the channel dimension when projecting from 1 to 32 channels; the remaining TCN blocks maintain a fixed channel size of 32. The dilation rate increases across blocks (1, 2, and 4), enabling the network to capture wider temporal dependencies while preserving causality by masking future information.

### C. Long-range Modeling with Mamba

Fig. 3(b) illustrates the long-range modeling using Mamba selective state space model. The TCN feature map (32 × 3600) is then passed to the long-range sequence modeling module based on the Mamba Selective State Space Model (Mamba-SSM). The Mamba layer (d_model = 32) captures global temporal dependencies using a state-space formulation optimized for long-context efficiency. A residual connection is applied, followed by Layer Normalization (LN), which stabilizes the temporal dynamics and preserves consistency across time steps. Next, the output is processed by a convolutional feed-forward network (FFN) consisting of two 1×3 Conv1D layers with ReLU activation to enhance non-linearity. This convolution-based FFN is particularly effective for ECG/RRI modeling because adjacent hidden states exhibit strong local temporal correlations [21]. Another residual connection and LN are applied, followed by dropout (0.2). The output maintains the same dimensionality (32 × 3600).

### D. Pre-AF Prediction Head

Fig. 3(c) shows the prediction head used for pre-AF classification. Global Average Pooling (GAP) and Global Max Pooling (GMP) are applied in parallel to aggregate long-range temporal patterns from encoded feature maps generated in Fig. 3(b). The outputs are concatenated to form a 64-dimensional vector. This vector is passed through two fully connected (FC) blocks, each consisting of an FC layer, BN, and ReLU activation. The first FC block reduces the dimensionality from 64 to 32, and the final FC layer maps to two output classes (pre-AF and NSR). A softmax activation produces the predicted probability of pre-AF or NSR.

### E. Training and Evaluation Protocol

To ensure subject-independent evaluation, we performed 5-fold subject-wise cross-validation and testing. AF subjects and NSR subjects were each partitioned into five mutually exclusive groups (20% each). For every fold, three groups (60%) were used for training, one group (20%) for validation, and one group (20%) for testing. No subject appeared in more than one split, ensuring strict subject-level independence. All training, validation, and testing analyses used the non-overlapping 1-h windows described in Section III-B. The performance metrics are reported as the mean values across the five folds.

For performance evaluation, we computed five key metrics: sensitivity (Sens.), specificity (Spec.), binary F1-score (F1.), area under the receiver operating characteristic curve (AUROC), and area under the precision-recall curve (AUPRC). Positive predictive value (PPV) and negative predictive value (NPV) were additionally calculated to assess performance at clinically relevant operating points. The binary F1-score and AUPRC were included specifically to assess pre-AF classification performance on the minority class, ensuring a robust evaluation despite dataset imbalance.

Hyperparameters were optimized using random search over the ranges listed in Table II. We explored batch size, learning rate, weight decay, TCN depth, TCN channel width, dropout rate, kernel size, the number of Mamba blocks, and sampling rate of RRI signals. The optimizer (AdamW) and the kernel initialization method (Kaiming/He uniform) were fixed and not included in the search. The best configuration, selected based on validation performance, used a batch size of 16, a learning rate of 0.0001, weight decay of 0.0001, three TCN blocks with 32 channels and kernel size 3, dropout of 0.2, one Mamba block, and 1-Hz sampling rate.

TABLE II
HYPERPARAMETER SEARCH SPACE AND SELECTED VALUES FOR AF-MAMBA MODEL

| Parameter | Search Space | Selected Value |
|---|---|---|
| Batch size | [16, 32, 64] | 16 |
| Learning rate | [1e-3, 1e-4, 1e-5] | 1e-4 |
| Weight decay | [1e-3, 1e-4, 1e-5] | 1e-4 |
| Dropout | [0.1, 0.2, 0.3] | 0.2 |
| Number of TCN blocks | [2, 3, 4, 5] | 3 |
| Channels in TCN blocks | [32, 64, 128] | 32 |
| Kernel size in TCN blocks | [3, 5, 7] | 3 |
| Number of Mamba blocks | [1, 2, 3] | 1 |
| Sampling rate | [1, 4] | 1 |
| Early stopping patience | [5, 10] | 10 |
| Optimizer | Fixed | AdamW |
| Kernel initializer | Fixed | Kaiming (He) |
| Epoch | Fixed | 1000 |

The best model was selected based on the lowest validation loss, using early stopping with a patience of 10 epochs. To align with clinical priorities, the decision threshold was set at the point corresponding to 0.9 sensitivity on the validation set and applied unchanged during testing. To further characterize the trade-off between sensitivity and false-alert burden, additional validation-derived operating points were evaluated. A false-positive alert was defined as an NSR window with an AF-Mamba prediction score exceeding the corresponding validation-derived threshold, and false-positive burden was reported as the number of alerts per 24 h of NSR monitoring. Model calibration was evaluated using pooled out-of-fold test predictions from the five folds using a reliability curve and the Brier score.

All experiments were conducted on Google Colab Pro using an NVIDIA A100 GPU (40 GB VRAM). To evaluate the suitability of the models for real-time implementation, we analyzed parameter count, computational complexity (FLOPs), and inference latency. We utilized the fvcore library to profile theoretical FLOPs. FLOPs for operations omitted by the profiler (Mamba, self-attention, RNN) were manually calculated and added to the reported totals. Inference latency was measured on the A100 GPU using CUDA events with a batch size of 1 (averaged over 100 runs) to simulate the real-time processing of a single patient's RRI segment.

### F. Additional Validation and Physiological Analyses

To assess generalizability across data sources and reduce the potential influence of dataset-origin confounding, we additionally performed paired cross-dataset holdout evaluation. Because each source database contributed only one class, conventional leave-one-dataset-out evaluation was not directly applicable. Instead, one AF dataset and one NSR dataset were simultaneously withheld for testing in each experiment. Subjects from both held-out datasets were excluded from model development, while training, validation, early stopping, and threshold selection were performed using only the remaining datasets. This procedure was repeated for all six possible AF–NSR dataset pairs.

We also evaluated whether model discrimination was primarily attributable to ectopic activity. Direct ectopic burden analysis was performed using LTAF, which provided both AF onset and reference PAC/PVC beat annotations together with sufficient continuous pre-AF recordings. PAC count, PVC count, and total ectopic burden were quantified within the same 1-h pre-AF windows used for model evaluation and were compared with non-overlapping 1-h windows from the MIT-BIH NSR and NSR-RR datasets. Using pooled out-of-fold predictions, windows were further stratified into low ($< 0.1\%$), moderate (0.1–0.5%), and high ($\geq 0.5\%$) ectopic-burden groups. AUROC and 95% confidence intervals were calculated within each group, with confidence intervals estimated using 1,000 subject-level bootstrap resamples.

To further determine whether AF-Mamba predictions were explained by conventional RR-interval irregularity measures, we performed an adjusted analysis using RMSSD, SD1/SD2,

SODP-Q1, SODP-CTM100, and PAS derived from the same 1-h windows. These five HRV measures were selected based on prior work demonstrating their strong predictive relevance for 1-h AF prediction [13]. These HRV measures were calculated from a 4-Hz RRI representation of the corresponding windows. Generalized estimating equation (GEE) models with a binomial distribution and logit link were fitted using pooled out-of-fold predictions. Repeated windows were clustered by subject using an independence working correlation structure with robust sandwich covariance estimates, and the cross-validation fold was included as a categorical covariate. The AF-Mamba prediction score was scaled such that the reported odds ratios corresponded to a 0.1 increase in model score.

Finally, we examined the temporal behavior of AF-Mamba prediction scores before AF onset. The risk-trajectory analysis included 37 subjects with 39 eligible AF episodes from IRIDIA-AF that had at least 24 h of continuous AF-free data before onset. Each episode was evaluated only using the model from the corresponding held-out cross-validation fold. Group-level trajectories were summarized with 95% confidence intervals estimated using 1,000 subject-level bootstrap resamples, and individual subject-level trajectories were also examined. For subjects with multiple eligible episodes, episode-level scores were averaged to obtain a single subject-level trajectory.

## V. Results

### A. Prediction Performance and Model Comparison

To evaluate the overall predictive performance of AF-Mamba, we compared the proposed model with existing RRI-based AF prediction models and representative general time-series architectures under the same subject-wise 5-fold evaluation protocol [13], [22], [23], [24], [25], [26], [15], [14]. As shown in Table III, AF-Mamba achieved a sensitivity of 0.889, specificity of 0.943, F1-score of 0.813, AUROC of 0.974, and AUPRC of 0.933. Among the existing AF prediction models, CNN-BiLSTM achieved competitive performance with an AUROC of 0.965 and AUPRC of 0.911, whereas the HRV-XGBoost and CNN-BiGRU models showed lower overall discrimination. Among the general time-series models, ResNet1D achieved the closest performance to AF-Mamba, with an AUROC of 0.972 and AUPRC of 0.905. Overall, AF-Mamba achieved the highest specificity, F1-score, AUROC, and AUPRC among the compared models, while maintaining comparable sensitivity.

We also performed an ablation analysis across different input lengths. As shown in Supplementary Table S1, AF-Mamba achieved the highest AUROC and AUPRC at each input duration, with performance progressively improving from 5 min to 60 min. For the 1-h input, AF-Mamba achieved an AUPRC of 0.933 and AUROC of 0.974, outperforming the alternative long-range modeling configurations in overall performance, although TCN-Transformer showed slightly higher sensitivity. These results support the effectiveness of the proposed TCN-Mamba architecture for long-duration RRI modeling.

TABLE III
Performance comparison of AF-Mamba with existing AF prediction and general time-series models

| Category | Model | Sens. | Spec. | F1. | AUROC | AUPRC |
|---|---|---|---|---|---|---|
| Existing AF prediction model | HRV-XGBoost [13] | 0.877 | 0.736 | 0.549 | 0.914 | 0.723 |
| | CNN-BiLSTM [22] | 0.885 | 0.920 | 0.770 | 0.965 | 0.911 |
| | CNN-BiGRU [23] | 0.891 | 0.646 | 0.493 | 0.911 | 0.810 |
| General time-series model | Inception Time [24] | 0.872 | 0.916 | 0.752 | 0.964 | 0.874 |
| | ResNet 1D [25] | 0.858 | 0.939 | 0.792 | 0.972 | 0.905 |
| | Vanilla Trans former [26] | 0.895 | 0.689 | 0.539 | 0.912 | 0.831 |
| | Vanilla Mamba [15] | 0.887 | 0.721 | 0.575 | 0.918 | 0.826 |
| | Vanilla TCN [14] | 0.885 | 0.576 | 0.483 | 0.817 | 0.605 |
| **Proposed** | **AF-Mamba (Ours)** | **0.889** | **0.943** | **0.813** | **0.974** | **0.933** |

Table IV compares the computational efficiency of AF-Mamba with the evaluated architectures. AF-Mamba achieved a favorable balance among model size, computational complexity, and inference latency, requiring 34.6 K parameters, 127.0 M FLOPs, and 2.75 ms per inference. Transformer-based architectures showed substantially higher computational costs for the 1-h input sequence. In particular, TCN-Transformer and the vanilla Transformer required 955.3 M and 1780.5 M FLOPs, respectively, compared with 127.0 M FLOPs for AF-Mamba, consistent with the increasing computational burden of self-attention for long input sequences. In contrast, deep convolutional architectures such as InceptionTime and ResNet1D required considerably larger model sizes, with 388.6 K and 10,367.3 K parameters, respectively. RNN-based architectures exhibited a different limitation, showing relatively high inference latency: TCN-BiLSTM, CNN-BiLSTM, and CNN-BiGRU required 6.47, 15.68, and 4.12 ms, respectively, compared with 2.75 ms for AF-Mamba, consistent with the sequential computation inherent in recurrent architectures. These findings indicate that AF-Mamba avoids the high computational cost of Transformer-based modeling, the large model size of deeper convolutional architectures, and the relatively high inference latency of recurrent architectures, while maintaining strong predictive performance. Overall, AF-Mamba provides a favorable performance–efficiency trade-off for long-duration RRI modeling.

TABLE IV
Efficiency comparison of the proposed AF-Mamba with other architectures

| Model | Params (K) | FLOPs (M) | Latency (ms) |
|---|---|---|---|
| **AF-Mamba (Ours)** | 34.6 | 127.0 | 2.75 |
| TCN-Transformer | 37.4 | 955.3 | 3.97 |
| TCN-BILSTM | 35.2 | 116.2 | 6.47 |
| TCN-Baseline | **18.3** | **57.22** | 1.65 |
| CNN-BiLSTM [22] | 304.6 | 1054.7 | 15.68 |
| CNN-BiGRU [23] | 152.1 | 541.0 | 4.12 |
| InceptionTime [24] | 388.6 | 1398.1 | 2.91 |
| ResNet1D [25] | 10367.3 | 2960.6 | 6.66 |
| Vanilla Transformer [26] | 33.9 | 1780.5 | **1.12** |
| Vanilla Mamba [15] | 73.8 | 309.1 | 3.31 |
| Vanilla TCN [14] | 66.7 | 250.3 | 3.85 |

### B. Cross-Dataset Generalization and Subject-Level Robustness

To further evaluate the generalizability of AF-Mamba beyond the datasets used for model development, we performed a paired cross-dataset holdout analysis in which one AF dataset and one NSR dataset were simultaneously excluded from training and validation and used exclusively for testing. As shown in Table V, AF-Mamba maintained discriminative performance across all six completely unseen AF–NSR dataset combinations, with AUROC values ranging from 0.797 to 0.991 and a mean AUROC of 0.897 ± 0.070.

However, performance varied across the held-out dataset combinations, particularly for threshold-dependent metrics. Sensitivity ranged from 0.250 to 0.938, while specificity ranged from 0.606 to 1.000. Lower F1-scores and AUPRCs were observed in several experiments in which NSR-RR was used as the unseen control dataset. Although threshold-dependent metrics varied across the held-out dataset combinations, AF-Mamba maintained meaningful discrimination, with AUROCs ranging from 0.797 to 0.991. However, the variability in sensitivity, specificity, and F1-score suggests that the decision threshold may require re-calibration across different data sources or populations.

We additionally examined whether the unequal number of eligible windows contributed by individual subjects affected the reported performance. Although AF subjects contributed an average of 1.60 windows per subject and NSR subjects contributed an average of 18.42 windows per subject, subject-level averaging yielded performance comparable to the primary window-level analysis. Specifically, the AUROC was 0.976 at the subject level compared with 0.974 at the window level, while the subject-level sensitivity, specificity, F1-score, and AUPRC were 0.900, 0.986, 0.944, and 0.991, respectively. These findings indicate that the model's discrimination was preserved when each subject contributed a single aggregated prediction. Detailed window distributions and subject-level results are provided in Supplementary Tables S2 and S3.

TABLE V
PAIRED CROSS-DATASET HOLDOUT EVALUATION OF AF-MAMBA

| Held-out AF dataset | Held-out NSR dataset | Sens. | Spec. | F1. | AUROC | AUPRC |
|---|---|---|---|---|---|---|
| IRIDIA-AF | MIT-BIH NSR | 0.818 | 0.997 | 0.896 | 0.991 | 0.991 |
| IRIDIA-AF | NSR RR | 0.500 | 0.894 | 0.512 | 0.836 | 0.532 |
| LTAF | MIT-BIH NSR | 0.250 | 1.000 | 0.400 | 0.910 | 0.727 |
| LTAF | NSR RR | 0.750 | 0.681 | 0.055 | 0.797 | 0.079 |
| MIT-BIH AF | MIT-BIH NSR | 0.875 | 1.000 | 0.933 | 0.976 | 0.940 |
| MIT-BIH AF | NSR RR | 0.938 | 0.606 | 0.073 | 0.873 | 0.121 |
| Mean ± SD | — | 0.688 ± 0.240 | 0.863 ± 0.161 | 0.478 ± 0.350 | 0.897 ± 0.070 | 0.565 ± 0.361 |

### C. Ectopy and RR-Irregularity Analyses

To examine whether AF-Mamba predictions were primarily driven by ectopic activity, we quantified ectopic burden in the 12 eligible pre-AF episodes from 9 LTAF subjects and compared them with non-overlapping 1-h windows from the NSR-RR and MIT-BIH NSR datasets. Pre-AF windows showed a higher median PAC count and total ectopic burden than the NSR windows. The median total ectopic burden was 0.119% [0.084–0.519] in the pre-AF windows, compared with 0.041% [0.000–0.137] in NSR-RR and 0.000% [0.000–0.000] in MIT-BIH NSR. In contrast, conventional RR-interval variability measures such as SDNN and RMSSD were not elevated in the pre-AF windows. Detailed ectopy and HRV characteristics are provided in Supplementary Table S5.

We next evaluated whether AF-Mamba discrimination was preserved across different levels of ectopic burden. As shown in Table VI, the AUROC was 0.842 (95% CI, 0.750–0.942) in the low-ectopy group, 0.854 (95% CI, 0.656–0.975) in the moderate-ectopy group, and 0.933 (95% CI, 0.807–1.000) in the high-ectopy group. The overall AUROC was 0.887 (95% CI, 0.784–0.975). Notably, discrimination remained substantial even in the low-ectopy group, suggesting that AF-Mamba performance was not limited to windows containing frequent PACs or PVCs.

To further assess whether AF-Mamba predictions were explained by RR-irregularity measures [13], we performed GEE analyses adjusting for RMSSD, SD1/SD2, SODP-Q1, SODP-CTM100, and PAS. Even after accounting for all five measures simultaneously, the AF-Mamba prediction score remained strongly associated with pre-AF status. Specifically, each 0.1 increase in the AF-Mamba score was associated with a 2.20-fold increase in the odds of a window being pre-AF rather than NSR (adjusted OR, 2.20; 95% CI, 1.92–2.52; $p < 0.001$). This indicates that AF-Mamba provided discriminative information beyond that captured by these conventional RR-irregularity measures. Similar results were obtained when each measure was adjusted for separately, as shown in Supplementary Table S6. Together with the preserved discrimination in the low-ectopy group, these findings suggest that AF-Mamba predictions were not fully explained by ectopic activity or the evaluated RR-irregularity measures.

TABLE VI
AF-MAMBA DISCRIMINATION ACROSS ECTOPIC BURDEN LEVELS

| Ectopic burden | Pre-AF windows | NSR windows | AUROC (95% CI) |
|---|---|---|---|
| Low (< 0.1%) | 4 | 1021 | 0.842 [0.750–0.942] |
| Moderate (0.1–0.5%) | 5 | 211 | 0.854 [0.656–0.975] |
| High (≥ 0.5%) | 3 | 94 | 0.933 [0.807–1.000] |
| Overall | 12 | 1326 | 0.887 [0.784–0.975] |

## D. Calibration and Operating Point Analysis

To assess the calibration of AF-Mamba predictions, we evaluated the agreement between predicted probabilities and observed pre-AF outcomes using pooled out-of-fold test predictions. As shown in Fig. 4, the reliability curve showed an overall increasing relationship between the predicted AF probability and the observed pre-AF frequency, indicating that higher model scores generally corresponded to a higher likelihood of pre-AF status. The pooled out-of-fold Brier score was 0.041, although some deviation from perfect calibration was observed at intermediate predicted probabilities.

We further evaluated AF-Mamba across multiple validation-derived operating points to characterize the trade-off between sensitivity and false-alert burden (Table VII). At the primary operating point targeting a validation sensitivity of 0.90, AF-Mamba achieved a test sensitivity of 0.889 ± 0.023, specificity of 0.943 ± 0.042, PPV of 0.760 ± 0.147, NPV of 0.978 ± 0.005, and F1-score of 0.813 ± 0.088, corresponding to 1.369 ± 0.998 false-positive alerts per 24 h of NSR monitoring. More restrictive false-alert criteria progressively reduced the number of false-positive alerts while increasing specificity and PPV, at the cost of lower sensitivity. For example, the operating point targeting ≤ 0.10 false-positive alerts/day yielded a sensitivity of 0.767 ± 0.070, specificity of 0.994 ± 0.005, PPV of 0.959 ± 0.036, and an observed false-positive burden of 0.148 ± 0.124 alerts/day. Conversely, targeting a sensitivity of ≥ 0.95 increased test sensitivity to 0.947 ± 0.032 and NPV to 0.988 ± 0.008, but reduced specificity to 0.822 ± 0.145 and PPV to 0.574 ± 0.227, with the false-positive burden increasing to 4.272 ± 3.492 alerts/day.

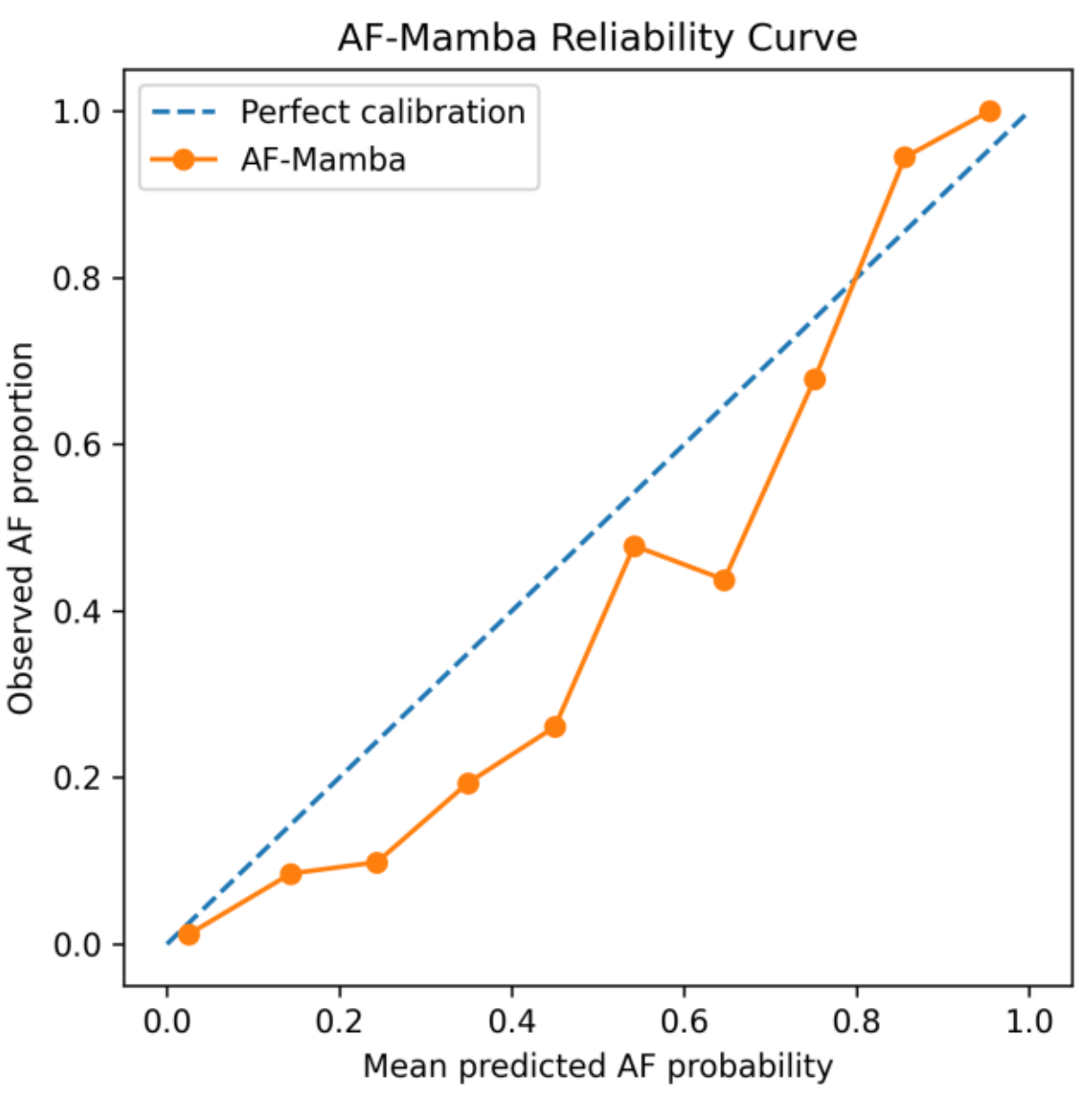

Fig. 4. Reliability curve of AF-Mamba based on pooled out-of-fold test predictions. The dashed diagonal represents perfect calibration. The Brier score was 0.041.

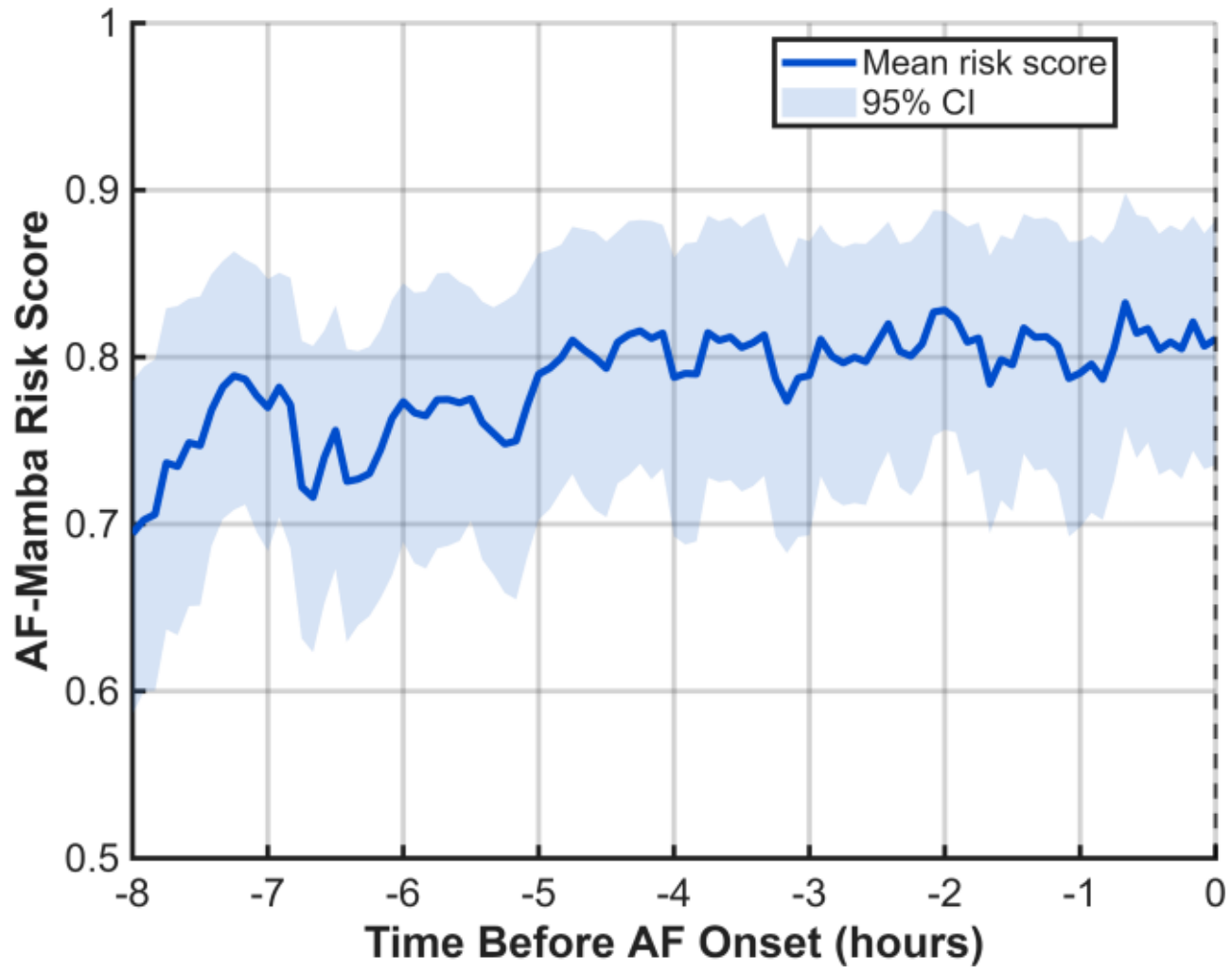

Fig. 5. AF-Mamba risk trajectory during the 8 h preceding AF onset. The solid line represents the mean AF-class prediction score, and the shaded region represents the 95% confidence interval estimated using 1,000 subject-level bootstrap resamples.

Overall, these results demonstrate a clear trade-off between detection sensitivity and false-alert burden. Therefore, the operating threshold can be adjusted according to the intended monitoring setting and the relative importance of minimizing missed pre-AF windows versus reducing false-positive alerts.

## E. Risk Trajectory Analysis

To examine the temporal changes of AF-Mamba predictions before AF onset, we analyzed 39 eligible AF episodes from 37 IRIDIA-AF subjects with at least 24 h of continuous AF-free data preceding onset. Each episode was evaluated using only the model from the corresponding held-out cross-validation fold. As shown in Fig. 5, the mean AF-Mamba prediction score was already elevated approximately 8 h before AF onset and increased modestly as onset approached, from approximately 0.70 to 0.80. The 95% confidence intervals were estimated using 1,000 subject-level bootstrap resamples.

Rather than exhibiting a sharp increase immediately before AF onset, the group-level trajectory remained persistently elevated throughout much of the 8-h pre-AF period. Individual subject trajectories also showed substantial heterogeneity and non-monotonic variation, including persistently elevated scores in some subjects and intermittent fluctuations in others (Supplementary Fig. S1). These findings suggest that the AF-Mamba score reflects a sustained state of elevated pre-AF susceptibility rather than a temporally precise estimate of AF onset exactly 1 h in advance.

TABLE VII
AF-Mamba Performance Across Validation-Derived Operating Points Representing Different Sensitivity and False-Alert Trade-Offs

| Validation criterion | Threshold | Sens. | Spec. | PPV | NPV | F1. | FP alerts/24-h NSR monitoring |
|---|---|---|---|---|---|---|---|
| FP/day ≤0.10 | 0.661 ± 0.188 | 0.767 ± 0.070 | 0.994 ± 0.005 | 0.959 ± 0.036 | 0.957 ± 0.011 | 0.851 ± 0.048 | 0.148 ± 0.124 |
| FP/day ≤0.25 | 0.601 ± 0.188 | 0.793 ± 0.064 | 0.989 ± 0.010 | 0.934 ± 0.061 | 0.962 ± 0.010 | 0.856 ± 0.053 | 0.254 ± 0.242 |
| FP/day ≤0.50 | 0.563 ± 0.186 | 0.813 ± 0.075 | 0.986 ± 0.010 | 0.913 ± 0.063 | 0.965 ± 0.013 | 0.857 ± 0.056 | 0.338 ± 0.241 |
| FP/day ≤1.00 | 0.384 ± 0.196 | 0.866 ± 0.074 | 0.957 ± 0.034 | 0.806 ± 0.134 | 0.974 ± 0.013 | 0.825 ± 0.074 | 1.036 ± 0.822 |
| **Sensitivity ≥0.90** | **0.299 ± 0.121** | **0.889 ± 0.023** | **0.943 ± 0.042** | **0.760 ± 0.147** | **0.978 ± 0.005** | **0.813 ± 0.088** | **1.369 ± 0.998** |
| Sensitivity ≥0.95 | 0.138 ± 0.106 | 0.947 ± 0.032 | 0.822 ± 0.145 | 0.574 ± 0.227 | 0.988 ± 0.008 | 0.685 ± 0.174 | 4.272 ± 3.492 |

## VI. Discussion

In this study, we developed the AF-Mamba architecture, a specialized deep learning framework designed for the early prediction of AF onset using long-term input sequences. The contributions of this work are threefold: 1) achieving strong predictive performance for 1-h AF prediction, 2) providing a favorable performance–efficiency trade-off for long-term RRI modeling, 3) demonstrating robustness across datasets, subjects, and potential physiological confounders.

A key advantage of AF-Mamba is its ability to model long RRI sequences without incurring the computational burden associated with many conventional deep learning architectures. Transformer-based models showed substantially greater computational complexity for the 1-h input, as shown in Table IV, reflecting the increasing cost of self-attention as sequence length increases. In contrast, deep convolutional architectures required considerably larger model sizes, whereas recurrent architectures exhibited relatively high inference latency because of their sequential computation. AF-Mamba does not achieve the lowest computational cost in every individual measure; rather, it provides a favorable balance among predictive performance, parameter count, computational complexity, and inference latency. In summary, the main technical advantage of AF-Mamba lies in its favorable performance–efficiency trade-off for long-duration RRI sequences.

Previous studies have adopted different strategies to address the difficulty of modeling long ECG/RRI sequences for AF onset prediction, as summarized in Supplementary Table S7. Earlier approaches primarily analyzed relatively short RRI segments (5-min) [11], [12], [27], whereas more recent methods incorporated longer-term information by combining predictions or representations from multiple shorter windows [13], [28]. In contrast, AF-Mamba directly processes the entire 1-h RRI sequence in an end-to-end framework, combining local feature extraction with efficient long-range temporal modeling. This enables long-duration RRI dynamics to be learned without intermediate short-window prediction or multi-stage aggregation, while maintaining a favorable performance–efficiency trade-off.

Beyond long-sequence modeling, the present study also evaluated the robustness of AF-Mamba across different data sources (Table V) and subject-level sampling patterns (Supplementary Table S3). The paired cross-dataset analysis showed that meaningful discrimination was retained when both the AF and NSR source datasets were unseen during model development. However, the variation in threshold-dependent performance across dataset combinations suggests that a decision threshold established in one dataset or population may not transfer directly to another and may require recalibration. The subject-level evaluation further showed that discrimination was maintained when each subject contributed a single aggregated prediction, suggesting that the observed performance was not primarily driven by differences in the number of eligible windows contributed by individual subjects. In addition, because the datasets used in this study were derived from Holter/ambulatory recordings, the generalizability of these findings to other clinical environments remains to be established.

We further examined whether AF-Mamba predictions could be explained primarily by ectopic activity or RR-irregularity measures [13]. Although ectopic activity was more frequent in the pre-AF windows (Supplementary Table S5), AF-Mamba maintained discrimination even among windows with low ectopic burden (Table VI). In addition, the AF-Mamba prediction score remained associated with pre-AF status after adjustment for multiple RR-irregularity measures (Supplementary Table S6). These findings suggest that AF-Mamba predictions are not fully explained by ectopic burden or the evaluated RR-irregularity features alone and that additional temporal information within the long-duration RRI sequence contributes to pre-AF discrimination.

Finally, the operating point and risk trajectory analyses provide important context for interpreting AF-Mamba predictions in continuous monitoring. The operating-point analysis demonstrated a clear trade-off between sensitivity and false-alert burden (Table VII), indicating that the appropriate decision threshold may depend on the intended monitoring setting. Importantly, the risk-trajectory analysis showed that prediction scores were already elevated several hours before AF onset and did not consistently increase as onset approached. Individual trajectories also showed substantial heterogeneity and non-monotonic variation. Therefore, the AF-Mamba score may be better interpreted as reflecting a sustained state of elevated pre-AF susceptibility rather than a temporally precise estimate that AF will occur exactly 1 h later. This finding suggests that future monitoring strategies may benefit from considering sustained elevations in AF-Mamba scores over time, rather than relying on a single-window prediction. The operating threshold could then be selected according to the intended monitoring setting.

## VII. Limitation

This study has several limitations. First, all analyses were retrospective and based on publicly available long-term Holter/ambulatory datasets. Although the paired cross-dataset holdout analysis showed that AF-Mamba retained discrimination when both AF and NSR source datasets were unseen during model development, AF and NSR windows originated from different source databases. Therefore, residual confounding related to acquisition conditions, patient characteristics, annotation procedures, or other dataset-specific factors cannot be fully excluded. In addition, the variability in threshold-dependent performance across held-out datasets indicates that calibration and operating thresholds may not transfer directly across populations or recording environments. Independent prospective validation and, if necessary, recalibration will therefore be required before clinical deployment.

Second, the present findings should primarily be interpreted in the context of ambulatory AF monitoring, as the datasets used in this study were derived from Holter/portable recordings. The generalizability of AF-Mamba to other clinical environments remains to be established, as differences in patient characteristics, recording equipment, signal quality, and monitoring conditions may affect model performance. Therefore, validation in prospective cohorts will be important before broader clinical application.

Third, the strict eligibility criteria may introduce selection bias, and the direct ectopy analysis was limited to 12 episodes from 9 LTAF subjects because suitable beat-level annotations were unavailable in the other AF datasets (Supplementary Table S4). Although the low-ectopy and HRV-adjusted analyses suggest that AF-Mamba predictions are not fully explained by ectopy or RR-irregularity measures, further validation in larger ectopy-annotated cohorts is needed.

Finally, AF-Mamba uses only RRI sequences as input. Although this representation is computationally efficient, it does not preserve morphological information contained in the raw ECG or incorporate potentially relevant clinical information. Future studies could investigate whether integrating ECG morphology or clinical variables with long-term RRI modeling further improves prediction. In addition, the present analysis was performed using quality-controlled RRI segments. Future studies should evaluate the robustness of AF-Mamba under noisier real-world recording conditions.

## VIII. Conclusion

This study proposed AF-Mamba, a deep learning architecture that integrates TCN-based local feature extraction with Mamba-based long-range sequence modeling for AF prediction 1 hour in advance using long-term RRI inputs. AF-Mamba achieved a sensitivity of 0.889, specificity of 0.943, F1-score of 0.813, AUROC of 0.974, and AUPRC of 0.933 under subject-wise evaluation using strictly non-overlapping 1-h windows. Importantly, AF-Mamba modeled the entire 1-h RRI sequence with only 34.6 K parameters and 127.0 M FLOPs, while requiring 2.75 ms per inference, demonstrating a favorable performance–efficiency trade-off for long-sequence modeling. Additional cross-dataset, subject-level, and physiological confounding analyses further supported the robustness of the proposed approach. Overall, these findings demonstrate the potential of AF-Mamba for efficient AF risk prediction 1 hour in advance using long-term RRI monitoring and motivate further prospective validation for real-world applications.

## Acknowledgment

This work expands the preliminary results presented at the 2025 IEEE International Conference on Body Sensor Networks (BSN) [29]. While the conference study used 30-min RRI inputs and evaluated multiple prediction horizons (0 to 2 h before), the present journal study focuses on direct end-to-end modeling of 1-h RRI sequences to predict AF 1 h before onset and includes substantially expanded datasets and validation analyses. The source code is publicly available at https://github.com/yongbin98/AF_Mamba.